 \documentclass[final,5p,times,twocolumn,authoryear]{elsarticle}

\usepackage{amssymb}
\usepackage{lipsum}
\usepackage{url}
\usepackage{amsmath}
\DeclareMathOperator*{\argmax}{arg\,max}

\usepackage{xcolor}
\usepackage{hyperref}
\usepackage{subfig}
\usepackage{graphicx}
\usepackage{booktabs}
\usepackage[T1]{fontenc}
\journal{Journal of Cycling and Micromobility Research}

\begin{document}

\begin{frontmatter}

%% Title, authors and addresses

%% use the tnoteref command within \title for footnotes;
%% use the tnotetext command for theassociated footnote;
%% use the fnref command within \author or \affiliation for footnotes;
%% use the fntext command for theassociated footnote;
%% use the corref command within \author for corresponding author footnotes;
%% use the cortext command for theassociated footnote;
%% use the ead command for the email address,
%% and the form \ead[url] for the home page:
%% \title{Title\tnoteref{label1}}
%% \tnotetext[label1]{}
%% \author{Name\corref{cor1}\fnref{label2}}
%% \ead{email address}
%% \ead[url]{home page}
%% \fntext[label2]{}
%% \cortext[cor1]{}
%% \affiliation{organization={},
%%            addressline={}, 
%%            city={},
%%            postcode={}, 
%%            state={},
%%            country={}}
%% \fntext[label3]{}

\title{A deep reinforcement learning solution to help reduce the cost in waiting time of securing a traffic light for cyclists}

\author[*,1]{Lucas Magnana}
\author[1]{Hervé Rivano}
\author[2]{Nicolas Chiabaut}
\affiliation[*]{organization={Corresponding author: lucas.magnana@gmail.com, 56 bd Niels Bohr},
            city={Villeurbanne},
            country={France}}
\affiliation[1]{organization={CITI, INSA Lyon-Inria, Université de Lyon},
            city={Villeurbanne},
            country={France}}
\affiliation[2]{organization={Département de la Haute-Savoie},
            city={Annecy},
            country={France}}

\begin{abstract}
Cyclists prefer to use infrastructures that separate them from motorized traffic. Using a traffic light to segregate car and bike flows, with the addition of bike-specific green phases, is a lightweight and cheap solution that can be deployed dynamically to assess the opportunity of a heavier infrastructure such as a separate bike lane. To compensate for the increased waiting time induced by these new phases, we introduce in this paper a deep reinforcement learning solution that adapts the green phase cycle of a traffic light to the traffic. Vehicle counter data are used to compare the DRL approach with the actuated traffic light control algorithm over whole days. Results show that DRL achieves better minimization of vehicle waiting time at every hours. Our DRL approach is also robust to moderate changes in bike traffic. The code of this paper is available at \url{https://github.com/LucasMagnana/A-DRL-solution-to-help-reduce-the-cost-in-waiting-time-of-securing-a-traffic-light-for-cyclists}.
\end{abstract}

%%Graphical abstract
%\begin{graphicalabstract}
%\includegraphics{grabs}
%\end{graphicalabstract}

%%Research highlights
%\begin{highlights}
%\item Research highlight 1
%\item Research highlight 2
%\end{highlights}

\begin{keyword}
%% keywords here, in the form: keyword \sep keyword, up to a maximum of 6 keywords
deep reinforcement learning \sep traffic light \sep cyclists \sep waiting time \sep 3DQN \sep vehicle counts

%% PACS codes here, in the form: \PACS code \sep code

%% MSC codes here, in the form: \MSC code \sep code
%% or \MSC[2008] code \sep code (2000 is the default)

\end{keyword}

\end{frontmatter}

%\tableofcontents

%% \linenumbers

%% main text

\section{Introduction}
\label{introduction}

Promoting cycling as a mode of transport is prevalent in urban policies worldwide, especially to reduce CO2 emissions of transportation \citep{mizdrak_potential_2019}. Cycling also saves residents time, money and improves their health \citep{oja_health_2011}. However, the cyclability of a city depends on space intensive infrastructures, such as bike lanes separated from the flow of cars, the feeling of safety being a major criterion for cyclists \citep{adam:hal-03552634,cervero_bike_and_ride_2013}. The development of innovative, space-saving infrastructures is therefore becoming a necessity. In particular, traffic lights could be used to separate the flows of bikes and cars, hence securing the former.
Each green phase of a traffic light allows certain vehicles that are not in conflict to pass through the intersection. Unfortunately, the green phase cycle is often independent on the traffic situation, making them unfit for bike isolation. However, recent advances in artificial intelligence for decision-making \citep{openai_dota_2019} and image recognition \citep{naranjo_torres_review_2020}, among others, can enable traffic lights to adapt to the traffic \citep{genders_open_source_2019}.

\subsection{Cyclists and traffic lights}

Cyclists are known not to always respect red lights. The proportion of cyclists observed running a red light varies from study to study, ranging from 40\% \citep{schleinitz_e_cyclists_2019} to 60\% \citep{richardson_investigating_2015}. \cite{johnson_why_2013} showed that in Australia, where people drive on the left, cyclists are more willing to infringe a red light when they want to turn left. The authors conclude that they cannot demonstrate that running a red light increases the likelihood of an accident. The red light infringements did not result in any risk to the safety of cyclists during their observations. \cite{hollingworth_risk_2015} showed a small increase in risk of accident-related injuries for cyclists infringing red lights. They note, however, that this increase could be caused by the generally riskier behavior of cyclists running red lights rather than actually running them. Traffic light-controlled intersections are nevertheless still dangerous places for cyclists. \cite{miranda_moreno_disaggregate_2011} studied the cyclist injury occurrence at traffic lights, and their results suggest that cyclists safety at traffic lights is significantly affected by cyclist volumes and traffic flows. The conflicts between motorized vehicles and cyclists, especially when it comes to right-turns, seem to significantly increase the risk of collisions in Montreal, Canada, where people drive on the right. Whether in Canada or Australia, dangerous behavior performed or experienced by cyclists increases in the case of trajectories that do not involve crossing other lanes \citep{johnson_why_2013, miranda_moreno_disaggregate_2011}. To address this issue in France, M12 signs indicate that cyclists may cross the intersection in specified directions when the light is red, with priority to vehicles with green lights.

The safety of cyclists at traffic lights is an issue. Some experiments try to help cyclists reach traffic lights when they are green. \cite{andres_co_riding_2019} created e-bikes designed to make cyclists catch a green wave. When the cyclist crosses the first green light, the e-bike adapts its assistance to make the cyclist go at the most adapted speed for the green wave. Similarly, \cite{frohlich_bikenow_2016} developed a smartphone application suggesting a range of speed allowing the cyclist to reach the next traffic light during a green phase. Other studies try to modify intersections for cyclists, which has the advantage of benefiting all cyclists and not just those with the right equipment. \cite{de_angelis_green_2019} asked cyclists to rate several interfaces at traffic lights, indicating whether cyclists are on time for a green wave. \cite{anagnostopoulos_cyclist_aware_2016} proposed traffic lights that prioritizes cyclists by detecting their smartphones. They however did not evaluate the impact of such a system on motorized traffic.

\subsection{DRL for traffic light control} \label{DRL_intro}

Deep reinforcement learning (DRL) has been used to adapt the behavior of traffic lights to the current traffic conditions and optimize the performance of the intersection. DRL is based on reinforcement learning (RL), an area of machine learning in which an agent develops its behavior through experience. The agent evolves in its environment and have the possibility to perform actions which modify it. At each step $t$, the agent receives the state of its environment $s_t \in S$ and chooses an action $a_t \in A$ with $S$ the set of all possible states and $A$ the set of possible actions per state. Once the action executed, the environment sends its new state $s_{t+1} \in S$ and a reward $r_t$ to the agent. The reward is a numerical value indicating how good or bad the action was. The goal of the agent is to develop a policy $\pi$ which maps an action to a state $\pi(s) = a$ as to maximize the cumulative reward $\sum_{t=0}^{T}{\gamma^t r_t}$ with $\gamma \in [0, 1)$ the discount-rate weighting the distant future events. DRL algorithms are RL algorithms in which the agent uses deep learning to make decisions (further explanations are given in Section \ref{DRL}). Some studies use DRL to modify a traffic light's pre-defined behavior. \cite{li_traffic_2016}, \cite{tan_deep_2019} as well as \cite{wei_intellilight_2018} used traffic lights with a static cycle and applied DRL to optimize the changing phase timing. The agent chooses whether the light switches to the next phase or remains at the current one, with a minimum time between two phase changes. \cite{genders_asynchronous_2019} used a traffic light with a static cycle and an initial duration for each phase. The DRL agent can increase or decrease the duration of a phase and has to find the optimum duration for each phase. 

Several studies used DRL to learn a dynamic cycle. In these, the DRL agent chooses at regular intervals which phase is the best. It selects not only the timing of phase changes, but also the order in which they take place. Some authors compare their DRL approaches to a deep learning approach \citep{genders_using_2016} or to other DRL approaches \citep{mousavi_traffic_2017}. \cite{wang_deep_2019} compared their DRL approach to one static and one dynamic traffic light control method on simulations with traffic demand evolving arbitrarily. \cite{genders_asynchronous_2019} compared their DRL approach to the same methods, but simulated peak hours to test its robustness in a more realistic setup.

\subsection{Positioning and contributions} \label{intersection} 

Cyclists are known to prefer infrastructure that allows them to ride away from cars \citep{caulfield_determining_2012, tilahun_trails_2007}. That's part of the reason some experimentation are planned in France to allow bikes to set off earlier in order to regain sufficient speed before the departure of other vehicles. The idea behind this work is to extend the latter concept by creating specific green phases for cyclists. This type of space-saving infrastructure would make it possible to separate bike and car flows just as well as conventional dedicated lanes, but at the expense of waiting time. Indeed, more green phases means a longer cycle, and therefore a longer waiting time between two green phases for all lanes. In this paper, we propose a DRL based green phase selection method, allowing the creation of specific green phases for cyclists with a limited impact on the waiting times at the intersection. The agent controls the order and the timing of phase changes. 
We use real life counts data %on a daily scale 
to compare our approach to %an 
existing traffic light control methods with realistic traffic. We hope that an infrastructure of this type with a sufficiently low impact on waiting time at the intersection would %help bring about 
foster
a modal shift %in favor of 
toward
cycling, thereby further increasing cyclists' safety \citep{elvik_non_linearity_2009}. The contributions of this paper can be summarized as :
\begin{itemize}
\item We propose a traffic light system that is safer for cyclists as it includes specific green phases for them.
\item We design a phase-change method using DRL to reduce the waiting time increase caused by such an infrastructure, using and improving existing designs.
\item We use real life counts data to test our approach on a daily scale.
\item We compare our approach to a dynamic one already deployed in order to demonstrate the relevance of using a DRL based solution.
\end{itemize}

\section{Deep reinforcement learning} \label{DRL}

To limit the increase in waiting time caused by the addition of green phases for cyclists, a DRL based phase-change method is proposed. This solution uses the Double Dueling Deep Q-Network (3DQN) algorithm, which is detailed in this section.

\subsection{Deep Q-Network}

In 2015, \cite{mnih_human_level_2015} developed an algorithm called Deep Q-Network (DQN) capable of learning human level policies. DQN is based on a reinforcement learning algorithm called Q-learning \citep{watkins_q_learning_1992}. A function $Q : S \times A \rightarrow \mathbb{R}$ calculates the quality of a state–action combination. Every time the agent chooses an action $a_t$, $Q(s_t, a_t)$ is updated using the Bellman equation. The final policy chooses the action with the best Q-value $\pi(s) = \max_aQ(s,a)$. In DQN, a deep neural network called Q-network approximates $Q$ and is noted $Q(s,a;\theta)$ where $\theta$ represents the parameters (i.e. the weights) of the neural network. The Q-network is trained by minimizing the loss function $L$ defined as : 
\[L(\theta) = (Y_t^{DQN} - Q(s, a; \theta))^2\]
\[Y_t^{DQN} = r_t + \gamma \max_{a'}{Q(s_{t+1}, a'; \theta')})\]
$Y_t^{DQN} $ is called the target value and $\theta'$ represents the parameters of the target network, a second neural network with the same architecture as the Q-network. The target network is only used to compute $Y_t$ and is updated towards the Q-network during training. 

\subsection{Double Deep Q-Network}

In 2016, \cite{van_hasselt_deep_2015} have shown that DQN has an overestimation bias for Q-values and proposed a solution inspired by the double Q-learning algorithm \citep{dqlearning} called Double Deep Q-network (DDQN). DDQN is very similar to DQN but calculates the target value a little differently:

\[Y_t^{DDQN} = r_t + \gamma Q(s_{t+1}, \argmax_{a'}{Q(s_{t+1}, a'; \theta)}; \theta')\]

In DQN, the action $a'$ used to calculate the target value is chosen and evaluated by the target network where in DDQN, the action is chosen by the Q-network and evaluated by the target network. This reduces the overestimation of Q-values, thus increasing the quality of the policies produced.

\subsection{Dueling Deep Q-Network} \label{dueling}

The advantage function is defined as $A(s,a) = Q(s,a)-V(s)$ where $V(s)$ represents the expected long term reward for being in the state $s$. From this equation, $Q$ can be decomposed as the sum of $V(s)$ and $A(s,a)$. The idea behind Dueling Deep Q-Network developed by \cite{pmlr_v48_wangf16} is to decompose the Q-network in two streams : $V(s; \theta)$ which approximates $V(s)$ and $A(s, a; \theta)$ which approximates $A(s,a)$. The Q-network and the target network are therefore defined as :

\[Q(s,a;\theta) = V(s; \theta) + (A(s, a; \theta) - \frac{1}{|A|}\sum_{a'}{A(s,a';\theta)})\]
\[Q(s,a;\theta') = V(s; \theta') + (A(s, a; \theta') - \frac{1}{|A|}\sum_{a'}{A(s,a';\theta')})\]

with $\theta$ the parameters of the Q-network and $\theta'$ the parameters of the target network. The mean of the approximations of advantages is subtracted from the approximations of $A(s, a)$ to increase learning stability and performance.

\subsection{Double Dueling Deep Q-Network (3DQN)}

The Double DQN and the Dueling DQN can be combined to obtain the Double Dueling Deep Q-Network (3DQN). 3DQN shows better learning stability and performance than DQN or DDQN in general. In 3DQN and all algorithms derived from DQN, the agent doesn't learn after every action it performs. Instead, it has a memory buffer in which it stores all the transitions $(s_t, a_t, s_{t+1}, r_t, d)$. This transition means that at the step $t$, the agent received $s_t$, chose the action $a_t$ which was rewarded by $r_t$ and made the environment in the state $s_{t+1}$. $d$ contains the information whether $s_{t+1}$ is a final state or not. The memory buffer has a finite size and if a new transition needs to be stored once it's full, the oldest is replaced by the new one. At each learning phase, the agent randomly fills a batch with transitions contained in the memory buffer and computes their mean loss. The mean loss is backpropagated to modify the weights of the Q-network. In our implementation, the weights of the target network are periodically replaced by the weights of the Q-network. Finally, an $\epsilon$ -greedy policy is used during training. When the agent needs to choose an action, it computes the Q-values using the Q-network. A random number $r \in [0, 1]$ is generated. If $r < \epsilon$, the action is chosen randomly. Otherwise, the action with the highest Q-value is chosen. $\epsilon$ is set to 1 at the beginning of the training and decreases as training progresses, in order to have a lot of exploration at the beginning and a lot of exploitation at the end.

\section{3DQN approach} \label{3DQN}
This section presents all the components the 3DQN agent will need to approximate an optimal policy, as well as how it is trained. First, the type of environment in which the agent will evolve is detailed. Then, the type of states that the environment will send to the agent is explained. The actions the agent will be able to perform in the environment, as well as the function rewarding them are defined. Finally, the training process and all the implementation choices made are explained.

\subsection{Environment} \label{environment}

As explained in Section \ref{intersection}, the traffic light has a green phase for each incoming car lanes but also for each incoming bike lanes. The environment in which the agent will evolve in is thus an intersection, made up of several intersecting axes, and controlled by a traffic light. For the sake of simplicity, each axis is assumed to be two-way, with a bike lane in each direction. If the car lanes on a given axis all have a green light at the same time (i.e. there are no specific green phases for turning cars), the light at an intersection of $n$ axes will have $2n$ different green phases. For example, the set of green lanes on an intersection of 2 axes will be $G = \{g^{ax_1}_{car}, g^{ax_1}_{bike}, g^{ax_2}_{car}, g^{ax_2}_{bike}\}$ with $g^{ax_x}_{t}$ meaning that the vehicle type $t$ on the axis $x$ has the green light. A graphical example of a two axes intersection is shown in Figure \ref{fig:state}, with $ax_1$ being the North-South (N-S) axis and $ax_2$ the East-West (E-W) axis.

\subsection{States} \label{states}

\begin{figure}
	\centering 
	\includegraphics[width=0.5\textwidth]{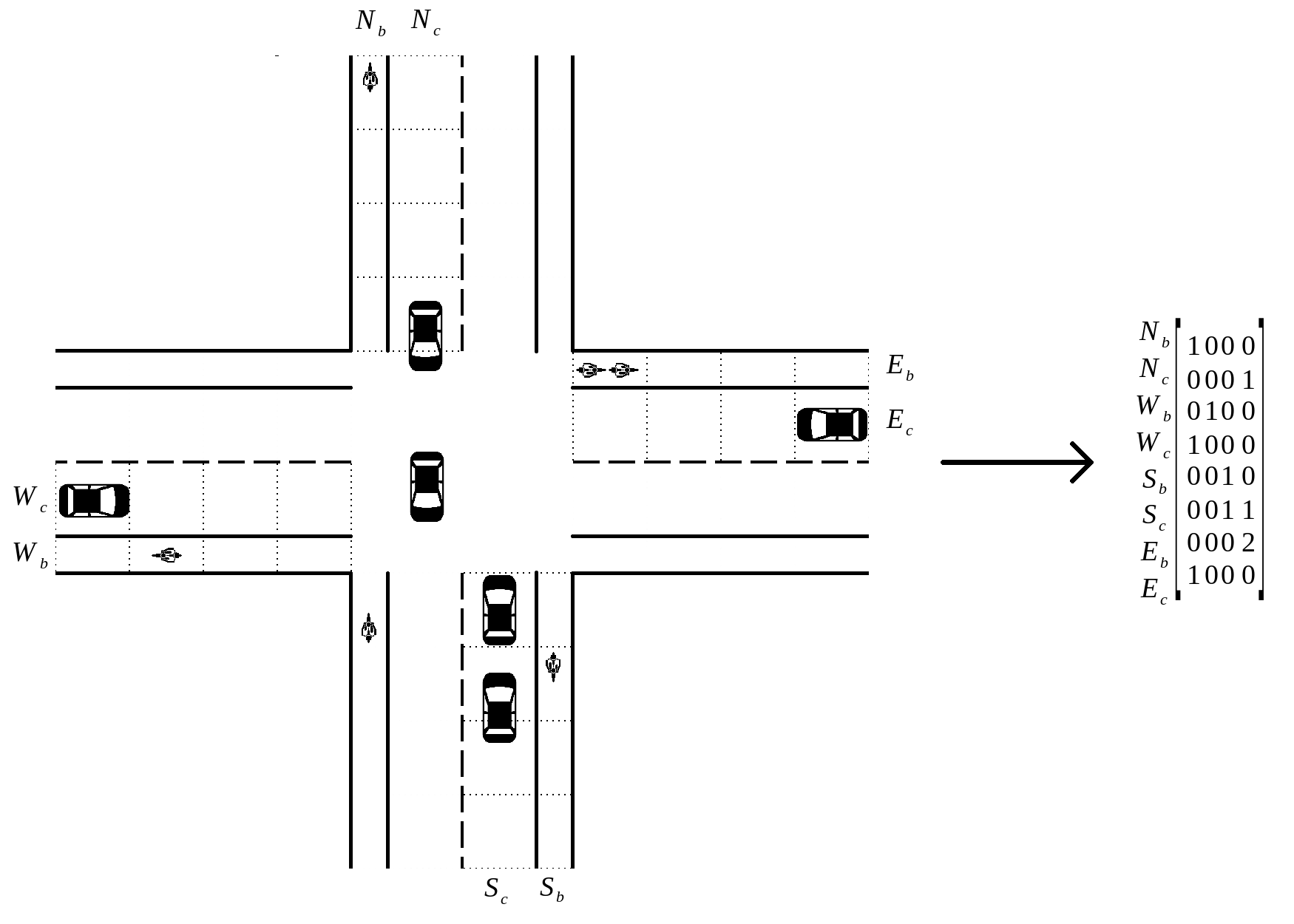}	
    \caption{Diagram showing the construction of the position matrix from an image of an intersection.} \label{fig:state}
\end{figure}

The states sent to the agent need to condense the useful information about the environment. We identified two type of information about the vehicles at the intersection that the agent needs to be informed of. First, the position of the vehicles. To make the best decision, the agent needs to know how many vehicles arrive at the intersection and on which lane they are. The second one is the speed. The agent needs to make the difference between the vehicles that are waiting and the one that are moving. The more vehicles waiting in a lane, the more it is necessary to change phase to let them pass. As in the work of \cite{liang_deep_2019}, the states are two matrixes of same dimensions. They are named respectively the position matrix and the speed matrix. The environment is divided in squares of 5-meters long. Only the squares belonging to the lanes arriving at the intersection are put in these matrices. The other ones can not contain useful information. The matrixes are thus smaller than those of \cite{liang_deep_2019}, as their dimensions are $N\times P$ with $N$ the number of lanes arriving at the intersection and $P$ the number of squares each lane contains. The position matrix contains the number of vehicles that are in each square, and the speed matrix contains the mean speed of the vehicles in each square. These matrixes could be reconstructed with cameras pointed at the lanes arriving at the intersection, and recent methods for estimating vehicle position and speed \cite{gunawan_detection_2019}. A graphical example of a position matrix is shown in Figure \ref{fig:state}.

\subsection{Actions} \label{actions}

The actions performed by the agent need to modify the behavior of the traffic light at an intersection. As it has been done several times \citep{genders_asynchronous_2019, mousavi_traffic_2017, wang_deep_2019}, the set of green phases $G$ is used as the action space (see Section \ref{environment}). Once a green phase start, 10s pass before the agent chooses the future green phase. If the chosen green phase is different from the actual one, a 4s orange light phase is triggered for the lanes that had green light until the decision. After this orange phase, or if the chosen green phase is the same as the actual one, a new 10s period is started before the agent chooses again. Waiting 10s after the start of a green phase avoids sudden changes that could surprise vehicles, and increases the stability of the agent's learning. 

\subsection{Rewards} \label{rewards}

\cite{genders_asynchronous_2019} used the same action space and developed a reward function working with it. The rewards they used are adapted as explained below.

\subsubsection{Reward function}

The reward function is defined as :

\[r_t = -(w_{b} + w_{c})^2\]

with $w_{b}$ and $w_{c}$ being respectively the number of waiting bikes and the number of waiting cars. A vehicle is considered to be waiting when its speed is less than 0.5 km/h. Note that the reward can only be negative, and that the more vehicles waiting, the more negative the reward. The agent must minimize the number of waiting vehicles in order to maximize the reward. 
The sum of $w_{b}$ and $w_{c}$ is squared to discriminate more strongly against bad decisions and facilitate the start of the training.

\subsubsection{Scaling factor}

However, the large negative values that the reward function can take,
especially at the start of training, may hamper the agent's convergence. \cite{genders_asynchronous_2019} coped with this issue by dividing the rewards by $r_{max}$, the biggest reward calculated. In our case, this normalization allowed the agent to converge, but has not led to the creation of effective policies. $r_{mean}$, the mean of all calculated rewards, is therefore used instead. All the calculated rewards as well as the number of actions performed by the agent during training are stored, and $r_{mean}$ is updated at the end of each training episodes (which are detailed in Section \ref{training}). Using $r_{mean}$ as a scaling factor incites the agent to perform actions that are better on average than those it has performed so far. As the agent improves, the average rewards will increase, pushing the agent ever further to become better. This allows the agent to finish training with a high-performance policy.

\subsection{Training} \label{training}

\subsubsection{Q-network architecture}

\begin{figure}
	\centering 
	\includegraphics[width=0.5\textwidth]{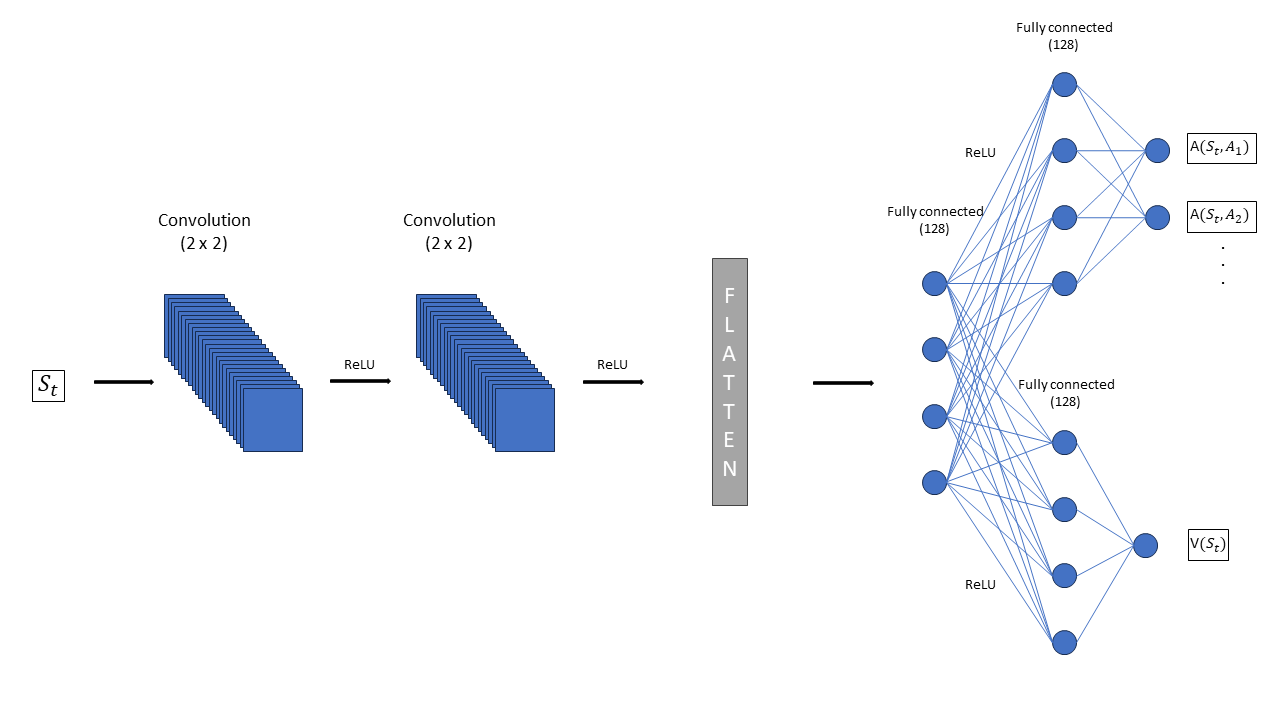}	
    \caption{Diagram showing the structure of the Q-network.}
    \label{fig:qnetwork}
\end{figure}

A state is made up of two matrixes that can be seen as a two channels image condensing the useful information of traffic at the intersection. Thus, the Q-network needs to be able to find patterns in an image in order to correctly process the states it receives. Convolutional layers extract features from images to lower dimensions without loosing their characteristics. A convolutional layer is composed of kernels, which are matrixes of small dimensions. Each kernel has its own weights in order to find a specific type of pattern in the image. The kernels slide along the image, and a multiplication is performed between their weights and the image values for each sub-area they cover. The Q-network is composed of two convolutional layers containing 16 kernels of dimension 2x2. These layers are followed by two fully connected layers of 128 nodes each. Then comes  two output layers for the value function and the advantage function (see Section \ref{dueling}).The ReLU activation function is used between all layers in order to provide non-linear properties to the Q-network. Figure \ref{fig:qnetwork} summarizes the architecture of the Q-network.

\subsubsection{Hyperparameters} \label{hyperparameters}

The values of the hyperparameters used during training as well as all the variables used in this paper are shown in Table \ref{table:variables} in Appendix \ref{a1}. The agent is driven in episodes. During each episode, vehicles can appear during 6 simulated hours, one step per second.
An episode ends when all vehicles have appeared and no vehicle remains in the simulation. A vehicle disappears once it has reached its destination, which is the end of one of the lanes leaving the intersection.
After acting $pt$ times, the agent learns at the end of each episode. Training stops when the agent has made $f$ actions. $\epsilon$ decreases linearly each time the agent chooses an action, and reaches its ending value at the $f^{th}$ action. Finally, the target network is updated by replacing its weights with those of the Q-network every $\upsilon$ actions performed by the agent.

% __________________________________________________>

\section{Experimental setup}

In this section, the simulated environment (cyclists, cars, and traffic lights) is presented. Then the synthesis of the traffic based on real count data is detailed. Finally, our performance evaluation methodology is explained.

\subsection{Simulated environment}

\begin{figure}
	\centering 
	\includegraphics[width=0.5\textwidth]{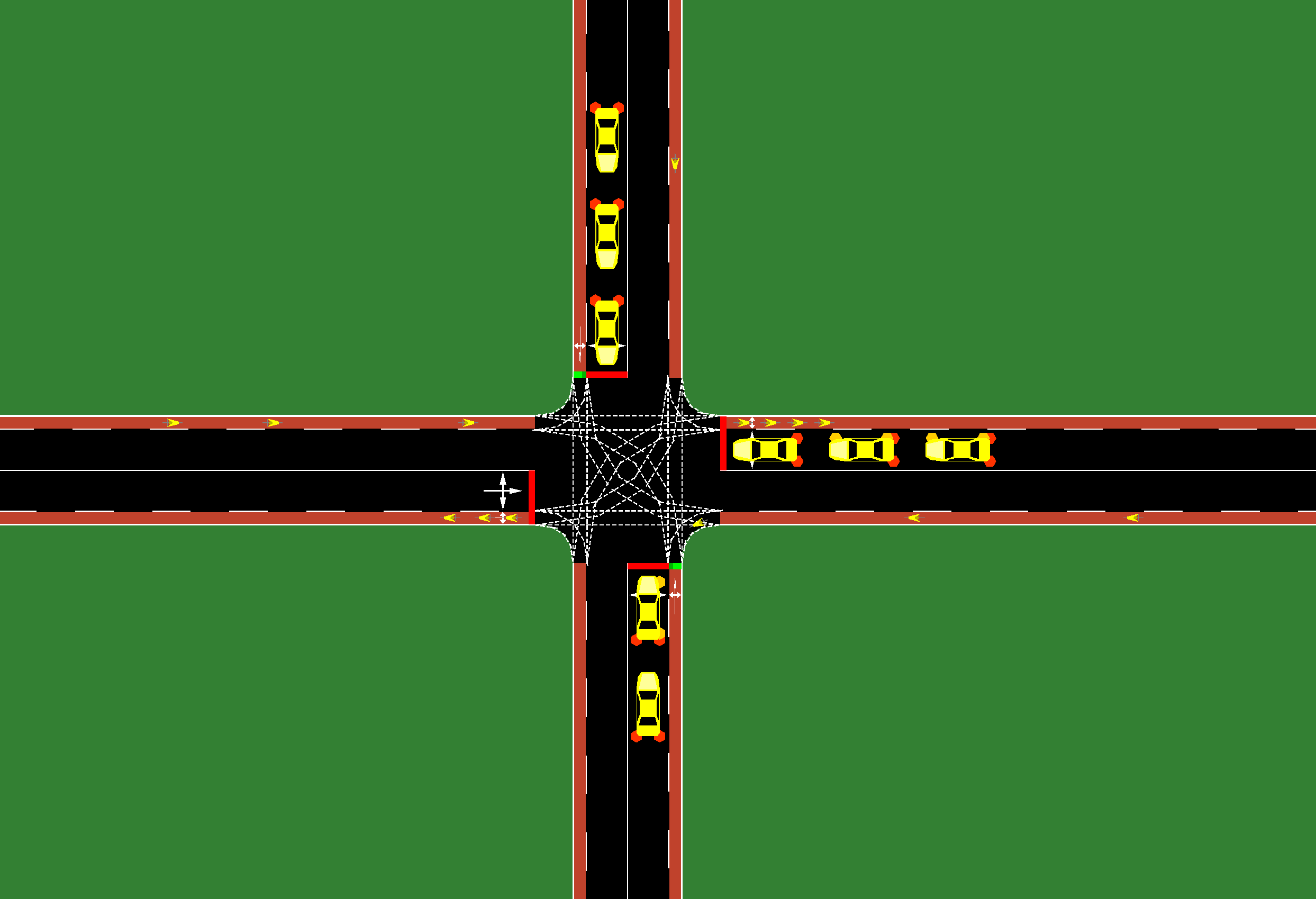}	
    \caption{Screenshot of the environment simulated by SUMO.} 
    \label{fig:intersection}
\end{figure}

SUMO (Simulation of Urban MObility) is used to simulate the environment. SUMO is a tool that allows different actors to interact on a road graph. The behavior of the different actors can be changed in real time. SUMO is commonly used in traffic light control studies using simulation. Figure \ref{fig:intersection} shows a screenshot of the environment. An intersection with two axes crossing (NS for North-South and EW for East-West) is managed by a traffic light. Each road arriving at the intersection is 150m long and has two lanes, one for bikes and one for cars. The traffic light has four green phases $G = \{g^{NS}_{car}, g^{NS}_{bike}, g^{EW}_{car}, g^{EW}_{bike}\}$. $g^{NS}_{bike}$ is activated on Figure \ref{fig:intersection}. The vehicles appear at the edge of a road and have the edge of another road as their destinations. When a vehicle is on green and wants to turn left, it must give way to oncoming traffic before crossing the intersection. This adds a waiting time that does not depend on the green phase of the traffic light. Moreover, when possible, vehicles tend to position themselves in the middle of the intersection when waiting to turn left to let vehicles behind them pass. Unfortunately, SUMO doesn't allow this behavior, making everyone wait behind it when a vehicle wanting to turn left is waiting to do so. This adds even more waiting time not dependent on the state of the traffic light. Vehicles are therefore prohibited from turning left. Vehicles have equal probabilities of having as destination one of the two roads they can access without turning left.

\subsection{Vehicle counts and traffic synthesis} \label{traffic_demand}

The city of Paris makes automatic vehicle counter data available on its open-data website \footnote{https://opendata.paris.fr/pages/home/}. Various temporal aggregations are available, from the year to the hour. Hourly aggregation is used here as it is the most precise. The data of two unidirectional car counters and one bidirectional bike counter are collected. The counters are located on boulevard Montparnasse and are close to each other. The boulevard Montparnasse is two-way, with two car lanes and a bike lane in each direction. The number of counted cars is halved, since our simulation has a single car lane in each direction. The data are from June 20, 2023, a Tuesday with good weather. 

\begin{figure}
     \centering
     \subfloat[][Hourly bike count.]{\includegraphics[width=0.5\textwidth]{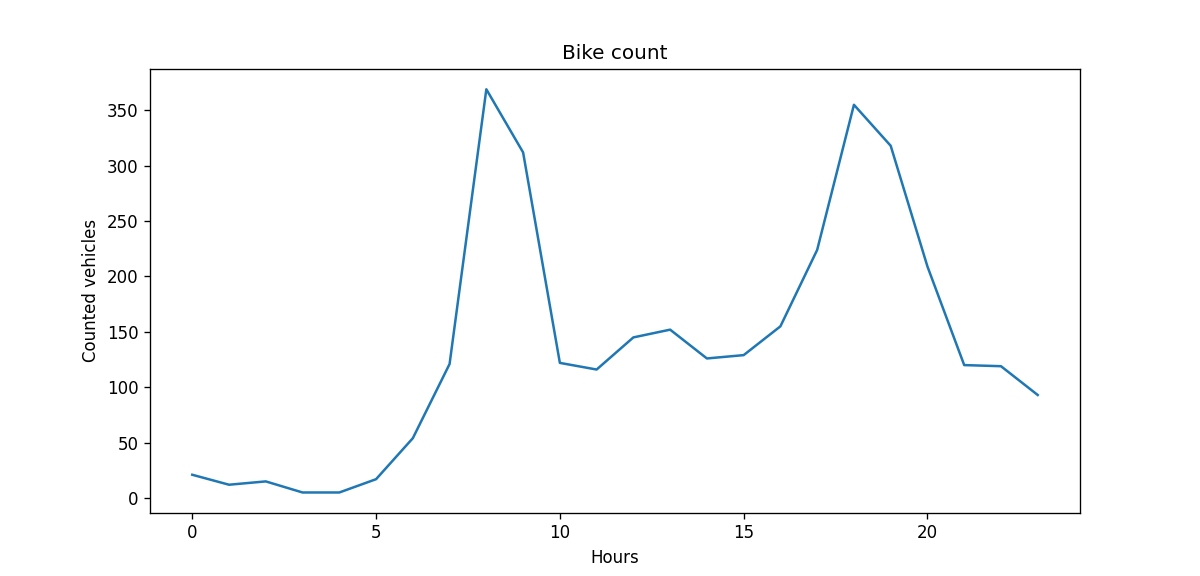}}
     \newline
     \subfloat[][Hourly car count.]{\includegraphics[width=0.5\textwidth]{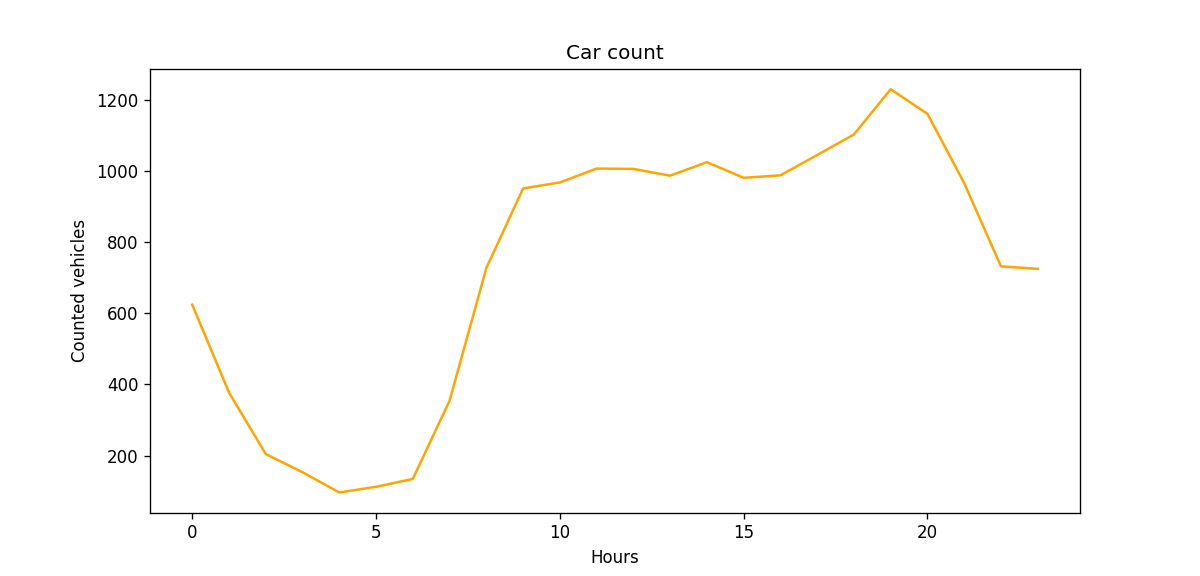}}
     \caption{Average number of vehicles counted per hour in both directions on boulevard Montparnasse on June 20, 2023.}
     \label{fig:counts}
\end{figure}

Figure \ref{fig:counts} shows the sum of the number of vehicles counted in both directions per hour. Differences between car and bike distribution are observable. There are hardly any bikes counted at night, a relatively stable number of bikes from 10:00 to 16:00 and two huge peaks at 08:00 and 18:00, the usual commuting hours. For the cars, the number decreases during all the night before increasing again at 06:00. The counted cars then reach a plateau that lasts until 17:00. Then comes a small increase with a peak at 19:00 before a decrease that will last until late at night. These differences between the vehicles distribution shows the importance of using real data on a daily scale. The 3DQN approach must be able to adapt its decisions to changes in both car and bike traffic in order to be efficient on a daily scale.

In order to simulate the traffic at the scale of each vehicle, we assume that the number of vehicle arriving at lane $l$ each second follows a Poisson process $p^l$. The intensity $\lambda_{p^l}(t)$ of this process at time $t$ is considered fixed during each hour and follows the aggregated count data :

\[\lambda_{p^l}(t) = \frac{c^{l}_{h(t)}}{3600}\] 

with $c^{l}_{h(t)}$ the number of vehicles (bikes or cars) counted at hour $h(t)$ on lane $l$. 

To summarize, the environment simulates the crossing of two Montparnasse boulevards during one day, each with only one car lane, and with traffic synthesized by extrapolating the hourly aggregation of real vehicles counts.

\subsection{Performance evaluation methodology} \label{compar_approach}

In our settings, the traffic is never saturated at the exit of the intersection. The performance of a solution is therefore the time spend by vehicles before they reach the intersection and get a green light. After training, whole days ($3600 \times 24 = 86400$ steps) are simulated and the mean waiting time of vehicles cars is calculated for each hour. The 3DQN approach is compared to the following traffic light control mechanisms.

\begin{itemize}
\item \textit{static unsecured} : The first approach compared, named \textit{unsecured} in the following, serves as a baseline to quantify the addition of waiting times when securing the traffic light for cyclists. This is a classic static traffic light, with only one green phase per axis. The bikes and the cars on the same axis cross the intersection at the same time.

\item \textit{static secured} : This approach is a naive one. It simply consists in preventing bikes from passing during the existing green phases and adding a green phase for bikes to all axes containing at least one bike lane. In our case, the traffic light has four green phases when using this approach, each lasting 40s. This behavior shows the huge increase in waiting time for all vehicles if the bike safety system is naively implemented.

\item \textit{actuated} : The \textit{static secured} approach serves mainly to demonstrate the importance of a dynamic phase-change method when implementing specific green phases for bikes. Comparing the 3DQN approach which is highly-dynamic only to a static approach would not be fair. Thus, \textit{actuated} is used. \textit{actuated} is a dynamic phase-change method commonly implemented in Germany \citep{brilon_priority_1994}. A traffic light in \textit{actuated} mode has vehicle detectors on each of its incoming lane, approximately 50m ahead. The traffic light has a $duration$ parameter, and each green phase has a minimum duration $minDur$ and a maximum duration $maxDur$. When a green phase start, the traffic light waits $minDur$ seconds before starting a counter of $duration$ seconds. Once the counter reaches zero, the traffic light switches to the next phase of its cycle. If one of the detector on the lanes with the green light detects a vehicle before the counter reaches zero, the counter is reset. If a green phase reaches a duration of $maxDur$, the traffic light switches to the next phase, regardless of the counter's status. In the implementation of \textit{actuated} used, all green phases have a $minDur$ of 10s, a $maxDur$ of 40s and the $duration$ parameter is set to 5s. \textit{actuated} is commonly used to test the performance of DRL approaches doing traffic light control \citep{wang_deep_2019, genders_asynchronous_2019, tan_deep_2019}.
\end{itemize}
% __________________________________________________>

\section{Results}

 The results are in two parts. The first ones are on an hourly scale for a simulation of one day. An initial simulation with a traffic light controlled by the 3DQN agent is carried out, with our random traffic synthesis. The times and places vehicles appear during this simulation are recorded. Three further simulations, one for each other control mechanisms, are then carried out with the traffic trace of the first simulation. This allows a comparison under perfectly identical conditions. The second part of the results is at a larger scale. The traffic demand of bikes is changed, and five different initial simulations are made for each bike traffic demand. The \textit{actuated} approach is then used for each initial simulation in the same way as explained above.

\subsection{Hourly results}  

\begin{figure}[hbtp]
     \centering
     \subfloat[][Hourly number of vehicles.\label{fig:hourly_results_num_v}]{\includegraphics[width=0.5\textwidth]{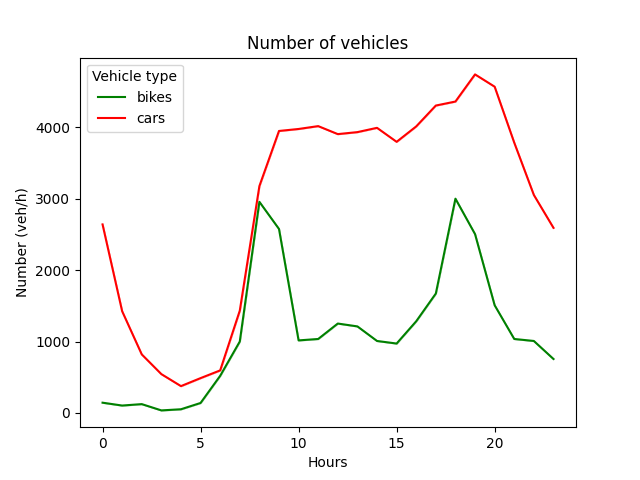}}
     \newline
     \subfloat[][Hourly mean waiting time.\label{fig:hourly_results_wt}]{\includegraphics[width=0.5\textwidth]{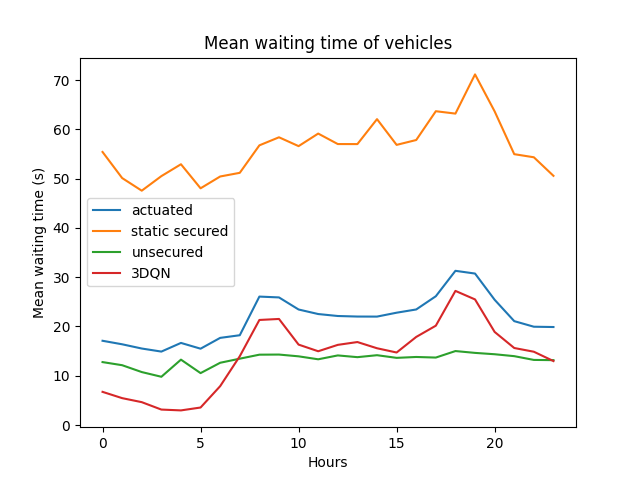}}
    \caption{Hourly number of vehicles and mean waiting time for a simulation of one day.} \label{fig:hourly_results}
\end{figure}

Figure \ref{fig:hourly_results} shows the results produced by a simulation of one day. The x-axis shows the hours of the day, from 0 to 23, and the y-axis shows the number of vehicles on Figure \ref{fig:hourly_results_num_v} and the mean waiting time of the vehicles on Figure \ref{fig:hourly_results_wt}. The vehicle distribution curves have the same shape as those shown in Figure \ref{fig:counts} without being perfectly identical. This is due to the randomness of the Poisson processes. 

As expected, the \textit{unsecured} approach does better than all the other ones. Unlike other curves, the \textit{unsecured}'s one is flat and changes little with traffic. The traffic is indeed not strong enough to saturate the road graph in this configuration. All vehicles waiting at a red phase are able to cross the intersection on the first green phase they are granted. Adding green phases dedicated to cyclists increases the occupancy rate of lanes since car lanes and bike lanes on the same axis are then emptied successively. 

The worst approach is the \textit{static secured} one. Doubling the duration of the traffic light's cycle results in an explosion of the waiting time, as each lane has to wait longer between two green phases. Reducing green phase durations to 20s increases the mean waiting time even further, with lanes becoming more saturated because they don't have enough time to empty during their green phases.

\textit{actuated} does much better than the \textit{static secured} approach whatever the time of the day, and handles the double car-bike peak at 19:00 with much greater ease. As the green phases duration are adapted to the traffic situation, the lanes are much less saturated. The green light time wasted for empty lanes is greatly reduced. 

The 3DQN approach does even better than \textit{actuated}, with a lower mean waiting time at every hour. The fewer vehicles there are, the better the performance of 3DQN. As the number of vehicles increases, the mean waiting time observed with 3DQN get close to that observed with \textit{actuated}. \textit{actuated} is logically less accurate when lanes are empty or almost empty. An \textit{actuated} traffic light follows a static cycle regardless of traffic conditions. 3DQN is able to adapt its cycle by selecting the lane that most needs to be green. This capability is important when vehicles are only present in certain lanes, which occurs more frequently when traffic is low. Thanks to its dynamic cycle, 3DQN performs even better than the \textit{unsecured} approach during nighttime hours when the number of vehicles is at its lowest. But when traffic increases, the phase change timing becomes more important than the cycle because vehicles are present in all lanes. Both \textit{actuated} and 3DQN having a dynamic phase change timing, the performance difference between the two approaches decreases when traffic increases.

On the scale of the day, adding specific green phases for cyclists multiplies the mean waiting time by 4.24 when using the \textit{static secured}, by 1.71 when using the \textit{actuated} approach and by 1.25 when using the 3DQN approach. Working with higher traffic would certainly allow us to reach the saturation rate of the \textit{unsecured} method, where it would be less effective, but this saturation would be all the more noticeable with the addition of green lanes for cyclists. This would further degrade the performance of the other approaches, and possibly even prevent the 3DQN agent's learning process.

\subsection{Robustness to changes in bike traffic}

\begin{figure}[hbtp]
     \centering
     \subfloat[][Number of vehicles.\label{fig:daily_results_sum_num_v}]{\includegraphics[width=0.5\textwidth]{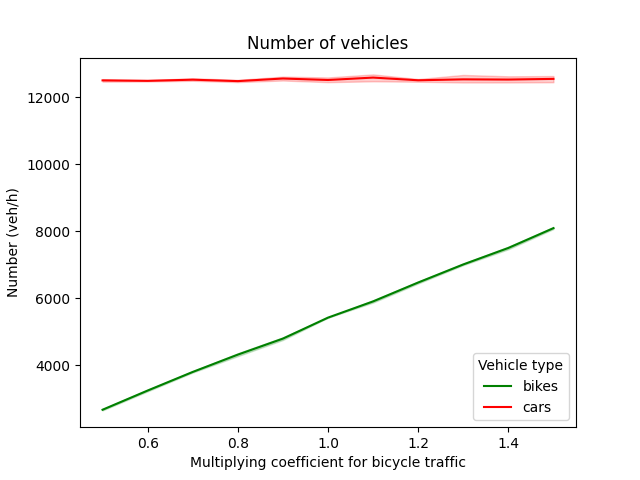}}
     \newline
     \subfloat[][Sum of waiting times.\label{fig:daily_results_sum_wt}]{\includegraphics[width=0.5\textwidth]{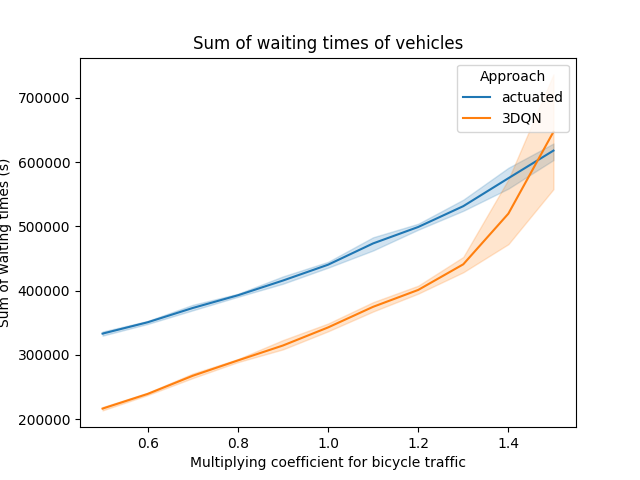}}
    \caption{Number of vehicles and sum of waiting times with respect to bike traffic changes (from 6h to 20h).}
    \label{fig:daily_results_sum}
\end{figure}

\begin{figure}[hbtp]
     \centering
     \subfloat[][Mean waiting times of bikes.\label{fig:daily_results_mean_wt_b}]{\includegraphics[width=0.5\textwidth]{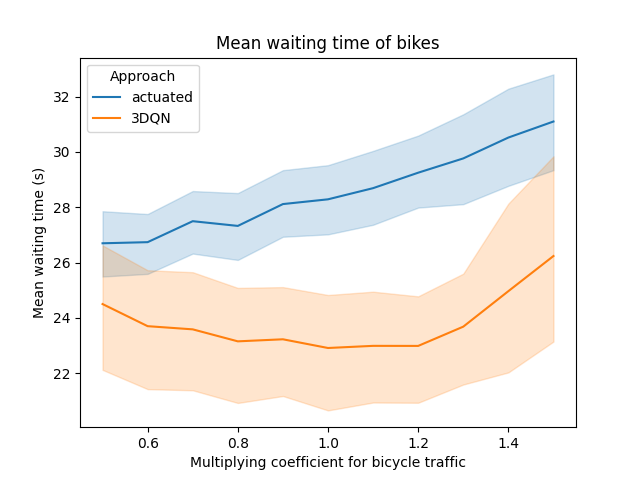}}
     \newline
     \subfloat[][Mean waiting times of cars.\label{fig:daily_results_mean_wt_c}]{\includegraphics[width=0.5\textwidth]{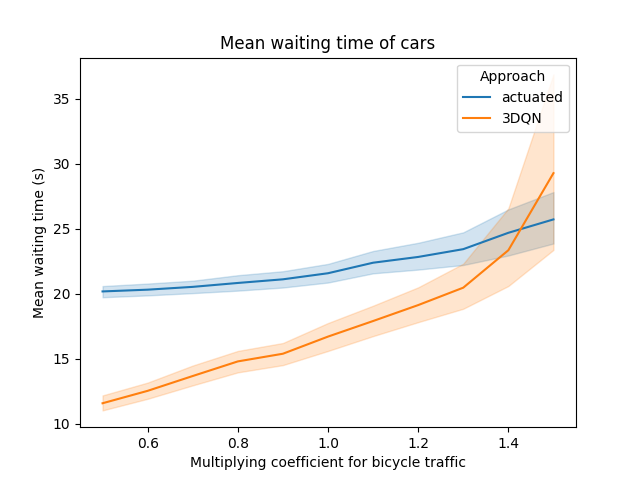}}
    \caption{Mean waiting times of vehicles with respect to bike traffic changes (from 6h to 20h).} \label{fig:daily_results_mean}
\end{figure}

The 3DQN approach limits the increase of the mean vehicle waiting times on a day with traffic similar to that used during training. However, the choice of whether to cycle is strongly correlated with the weather. As the traffic demands are calculated on the basis of data from a sunny day, the number of cyclists is likely to be lower on bad weather days. On the other hand, the aim of the secured intersection is to provide safe passage for cyclists, and the deployment of such infrastructure could attract cyclists, leading to an increase in bike traffic. The robustness of a 3DQN approach to changes in bike traffic therefore appears to be an important point to check. A multiplying coefficient which goes from 0.5 to 1.5 in steps of 0.1 is set. The number of bikes counted each hour is multiplied by this coefficient. That varies the bike traffic linearly from 50\% to 150\% of the observed one in the count data. Five days are simulated for each new bike traffic. \textit{actuated} being the best comparison approach on a secured traffic light, it is used to evaluate the performance of 3DQN. Since nighttime hours are not very relevant due to the absence of traffic, the results shown in Figure \ref{fig:daily_results_sum} and \ref{fig:daily_results_mean} focus on the hours between 6h and 20h.

Figure \ref{fig:daily_results_sum} shows the number of vehicles (\ref{fig:daily_results_sum_num_v}) and the sum of all the waiting times (\ref{fig:daily_results_sum_wt}) for each multiplying coefficient. Logically, the number of cars per simulation is stable and the number of bikes is increasing linearly. The sum of vehicle waiting times generated by 3DQN starts out lower than the one generated by \textit{actuated}. The two increase progressively, finally coming together when the coefficient reaches 1.5. When the coefficient is 1.5, 3DQN makes vehicles wait longer than \textit{actuated}. This is consistent with hour-by-hour observations. The fewer vehicles there are at the intersection, the higher the probability that \textit{actuated} will leave an empty lane green due to its fixed cycle. This logically favors 3DQN, which is by nature more dynamic. It still shows that 3DQN is able to adapt to a decrease in bike traffic without any significant impact on its decision-making performance. This is probably due to the high variations in both car and bike traffic that the agent faces during training, with nighttime hours having very low traffic levels. 3DQN's performance is also fairly stable as the number of bikes increases. The sum of waiting times logically increases, as more bikes means more green light time is needed to clear the bike lanes, which impacts the waiting times of all vehicles at the intersection. The difference in performance between the two approaches is stable until the coefficient reaches 1.3. When the multiplying coefficient is greater than 1.3, the difference in performance between the two approaches diminishes. The first reason for this is that, as explained above, the more vehicles there are, the better the \textit{actuated} performance. But the discrepancy in performance between simulations using the 3DQN approach also increases significantly when the multiplying coefficient is greater than 1.3. This is probably due to less relevant decisions made by the 3DQN agent. The further the bike traffic moves away from the one used during training, the greater the probability that the agent will receive a state it is not accustomed to handling. This situation may lead the agent not to make the best possible decision, resulting in more saturated lanes that it is no longer able to manage properly. When the coefficient reaches 1.5, 3DQN makes vehicles wait longer than \textit{actuated}. The 3DQN agent therefore appears to be robust with bike traffic ranging from 50\% to 140\% of the traffic used during training.

To go into more details, Figure \ref{fig:daily_results_mean} shows the mean waiting time for bikes (\ref{fig:daily_results_mean_wt_b}) and cars (\ref{fig:daily_results_mean_wt_c}) with respect to the multiplying coefficient. For \textit{actuated}, mean waiting times evolve in a fairly similar way for both types of vehicle, with a slight linear increase. 3DQN is different. For cars, there is a much more pronounced linear increase when the coefficient is lower than 1.3. With a coefficient greater than 1.3, the mean waiting time of cars increases more abruptly and exceeds that of \textit{actuated} when the multiplier coefficient reaches 1.5. On the other hand, the mean waiting time for bikes is stable and even decreases until the coefficient reaches 1.2, before rising slightly. This is a surprise, as we expected the mean waiting times for the two types of vehicles to evolve in the same way as the sum of the waiting times in Figure \ref{fig:daily_results_sum}. Instead, the mean waiting time for cars increases more sharply, allowing bikes to wait less. The agent seems to wait for a lane to reach a certain occupancy rate before turning it green, thus favoring cyclists in the experiment. Indeed, as the number of cyclists in the simulations increases, the bike lanes reach this occupancy rate more quickly, prompting the agent to turn them green more often. As a result, the waiting time for bikes does not increase despite their higher numbers, but cars are given the green light less often. That's why the mean waiting time for cars increases this way. The sum of mean waiting times observed with 3DQN ends up exceeding that of \textit{actuated}, because by giving preference to bikes in this way, the mean waiting time for cars ends up being too great for the agent to be as efficient as \textit{actuated}.

 % __________________________________________________>

 \section{Conclusion}

 In this paper, we proposed a traffic light allowing cyclists to cross an intersection safely during dedicated green phases. Results show that adapting the behavior of a traffic light to the traffic situation using DRL can reduce the cost in waiting time induced by securing the passage of cyclists with dedicated green phases. A 3DQN agent is capable of controlling this type of secure traffic light with different levels of traffic, enabling it to absorb fluctuations in traffic on a typical day. Performance remains relatively stable with moderate deviations from training conditions in terms of bike traffic. However, even though real count data are used to simulate traffic, it is modelled with Poisson processes, which is an unrealistic simplification of traffic demand. In the same spirit of realism, the vehicles in the simulations use SUMO's default behavior, which perfectly respects the rules of the road. Experiments with traffic demand and individual behaviors closer to reality must be carried out before such an infrastructure can be deployed. If these experiments prove conclusive, an interesting extension of our work would be to observe the distribution of vehicles exiting the intersection and to train another DRL agent with these distributions, with the aim of creating DRL driven green waves along a path with several intersections. We would also like to point out that an agent has been trained using another type of (policy-based) DRL algorithm called Proximal Policy Optimization (PPO). Although this agent has converged, its final policy performs less well than that of the 3DQN agent, but we don't know whether this is due to the nature of PPO or to our implementation. We are making available the code containing both algorithms for future works.

 % __________________________________________________>

 \section*{Competing interest statement}

 The authors declare that they have no competing interests that relate to the research described in this paper.

\appendix

\section{Variables used} \label{a1}

\begin{table}[ht]
\begin{tabular}{lcccc} \toprule
Section & Name & Value \\ \midrule
\ref{DRL_intro} & $s_t$ & Section \ref{states} \\
& $a_t$ & Section \ref{actions} \\
& $r_t$ & Section \ref{rewards} \\

\midrule
\ref{hyperparameters} & Memory buffer size & 25000 \\
& Batch size & 128 \\
& Starting $\epsilon$ & 1 \\
& Ending $\epsilon$ & 0.01 \\
& Pre-training acts $pt$ & 10000 \\
& Final action $f$ & 1500000 \\
& Discount-rate $\gamma$ & 0.99 \\
& Target update $\upsilon$ & 7500 \\
& Learning rate & 0.001 \\ 

\midrule
\ref{traffic_demand} & $\lambda_p$ & Section \ref{traffic_demand} \\
 
 \midrule
\ref{compar_approach} & $minDur$ & 10 \\
 & $maxDur$ & 40 \\
 & $duration$ & 5 \\ 
 \bottomrule
\end{tabular}
\caption{Table summarizing the variables used.}
\label{table:variables}
\end{table}

%% If you have bibdatabase file and want bibtex to generate the
%% bibitems, please use
%%
\bibliographystyle{elsarticle-harv} 
\bibliography{example}

%% else use the following coding to input the bibitems directly in the
%% TeX file.

%%\begin{thebibliography}{00}

%% \bibitem[Author(year)]{label}
%% For example:

%% \bibitem[Aladro et al.(2015)]{Aladro15} Aladro, R., Martín, S., Riquelme, D., et al. 2015, \aas, 579, A101

%%\end{thebibliography}

\end{document}